\documentclass[12pt,conference]{IEEEtran}

\usepackage{cite}
\usepackage{amsmath,amssymb,amsfonts}
\usepackage{tabularx}
\usepackage{algorithmic}
\usepackage{graphicx}
\usepackage{textcomp}
\usepackage{xcolor}
\def\BibTeX{{\rm B\kern-.05em{\sc i\kern-.025em b}\kern-.08em
    T\kern-.1667em\lower.7ex\hbox{E}\kern-.125emX}}
\begin{document}

\title{Cloud Removal from Satellite Images}

\author{\IEEEauthorblockN{Rutvik Chauhan}
\IEEEauthorblockA{\textit{Computing Science(Multimedia)} \\
\textit{University of Alberta}\\
Edmonton, Canada \\
rchauha1@ualberta.ca}
\and
\IEEEauthorblockN{Antarpuneet Singh}
\IEEEauthorblockA{\textit{Computing Science(Multimedia)} \\
\textit{University of Alberta}\\
Edmonton, Canada \\
antarpun@ualberta.ca}
\and
\IEEEauthorblockN{Sujoy Saha}
\IEEEauthorblockA{\textit{Computing Science(Multimedia)} \\
\textit{University of Alberta}\\
Edmonton, Canada \\
sujoy1@ualberta.ca}
}

\maketitle

\begin{abstract}
In this report, we have analyzed available cloud detection technique using sentinel hub. We have also implemented spatial attention generative adversarial network and improved quality of generated image compared to previous solution \cite{b7}. 
\end{abstract}

\section{Introduction}
Observing and monitoring the Earth's surface from space is essential to handle climate change, natural resources management, and disaster mitigation.  Remote Sensing (RS) imagery is critical to perform these challenging operations. However, clouds are one of the common factors which can create a complete or partial obstacle in the scene. 

Satellites examine the Earth’s surface across visible and infrared bands to capture images. The resulting multispectral image has more channels compared to an RGB image, corresponding to a different wavelength. Approximately half of the Earth’s surface is covered in opaque clouds with an additional 20\% being obstructed by cirrus or thin clouds. In some places, the clouds can cover the Area of Interest (AOI) for several months in a row. An abundance of research has already been done to remove the clouds and get the accurate pixels in the satellite images. The approaches include Traditional and ML-based solutions.

\section{Motivation}
In today's world, Satellite images depict the necessary information about the Earth's surface. This information can be crucial in various studies like disaster management and prevention. Cloud can impede the visibility of these image scenes. Thus, it is necessary to identify and replace the clouds with accurate pixels to make the images efficient for further general or ML-based studies. 
Researchers have done several types of research to detect and replace the clouds from satellite images. These studies can be categorized into the Traditional approaches and advanced ML-based approaches. 

Traditional approaches can be classified into three main groups: 1) Multi-Spectral, 2) Multi-Temporal, and 3) Inpainting. Multi-Spectral approaches can be applied in partially cloud-affected images. Multi-Temporal approaches use previously taken data to replace the cloud in the image. Lastly, Inpainting approaches can use cloud-detection algorithms to identify the cloudy part of an image so as to reconstruct it using the other non-cloudy ones.

In parallel with the traditional approaches, advanced machine learning-based algorithms like deep Convolutional Neural Networks (CNNs), Generative Adversarial Networks (GANs) have proven their efficiency to restore the missing information. However, the performance of the model can further be improved by ensembling more neural network algorithms.

\section{Related Work}
For cloud removal from satellite images, many different approaches have been proposed over the years. They can be broadly classified into the following categories: 
{\bf Traditional approaches} which involve detection of clouds from images and using image inpainting techniques to fill the cloudy areas with the help of cloud-free images taken over different time intervals. In \cite{b1}, a similar approach for removal of clouds from satellite images is presented. It involves usage of a threshold value to detect cloudy pixels from satellite images and replacing those pixels with the data from another images of the same location. Due to variations of patterns observed in clouds, selection of one definite threshold is a challenge.

{\bf Generative Adversarial Network(GAN) based approaches:} In \cite{b2}, authors have proposed a cyclic generative adversarial network, that eliminates the need of a paired training dataset and Synthetic Aperture Radar (SAR) images. They introduced a set of generator and discriminator which compete with each other until the generator can produce realistic cloud free images. Due to lack of paired cloudy-cloud free images they didn't have any quantitative results.

\textbf {Combination of multiple models based on deep learning techniques:} In \cite{b3}, authors have introduced two new paired datasets which they have created using real-world images available from European Space Agency satellite Sentinel-2 and a novel Spatiotemporal Generator Networks (STGAN) which utilizes RGB and IR data to determine relationships across multiple images over a location. 

In \cite{b4}, SpaceEye, a first of its kind model which utilizes both multi-modal and multi-temporal satellite images. SpaceEye takes multi spectral and SAR images as input, and uses a combination of a coarse and refinement network to patch the missing cloud pixels in an image. The predictions generated by this model are validated using a discriminator network which further improves quality of SpaceEye. Authors also stated the applications of SpaceEye by conducting case-studies.

In \cite{b5}, authors have been able to generate the entire image instead of only cloudy pixels by using a hierarchical combination of 3 neural networks - conditional Generative Adversarial Network (cGAN), Convolutional Long Short-Term Memory (ConvLSTM) and U-shaped Convolutional Neural Network. This method has shown encouraging results when compared with state-of-the-art techniques.

\section{Dataset}
We have the RICE dataset, which was proposed in \cite{b6}. The RICE1 dataset contains 500 data samples with each sample having a cloudy image and a cloudless image under 512×512 resolution. The dataset is collected by Google Earth, and the cloudy/cloudless images are obtained by setting the cloud layer whether to display. Figure 1 shows some examples of RICE1 dataset.
\begin{figure}[htp]
    \centering
    \includegraphics[width=8cm]{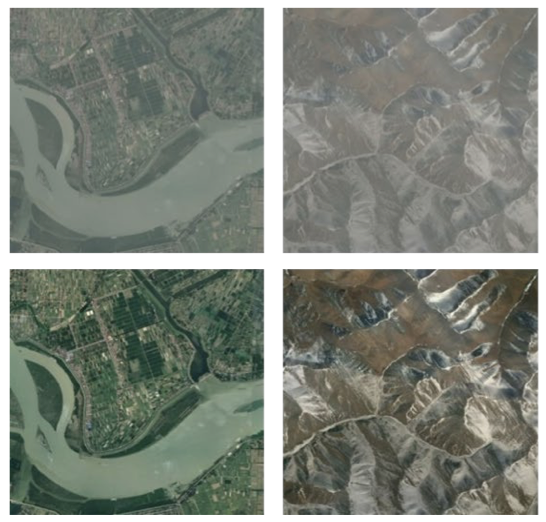}
    \caption{Upper row shows cloudy images and lower row shows cloud-free images from RICE1 dataset}
\end{figure}

\section{Implementation}
\subsection{Cloud detection using Sentinelhub and s2cloudless}
We have implemented a process for detecting clouds from sentinelheub2 satellite images using Sentinelhub and s2cloudless. The method receives dynamic satellite data from the Sentinel2 server using API requests. Then it generates the cloud mask of the particular satellite image using S2PixelCloudDetector. We used 10 out of 13 bands to create the cloud mask.

\begin{figure}[htp]
    \centering
    \includegraphics[width=5cm]{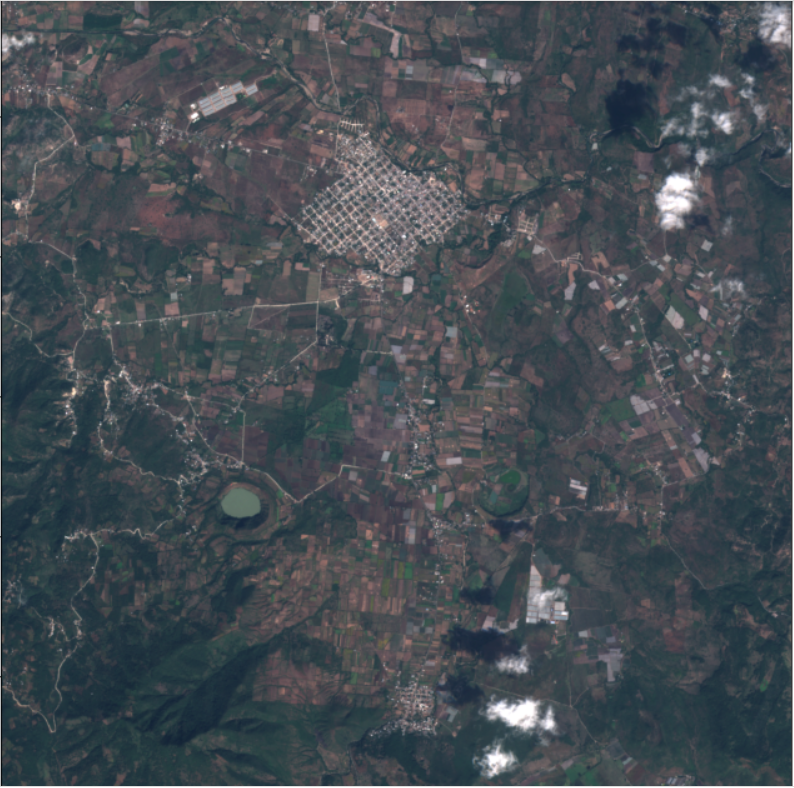}
    \caption{Sentinel2 image}
\end{figure}

\begin{figure}[htp]
    \centering
    \includegraphics[width=5cm]{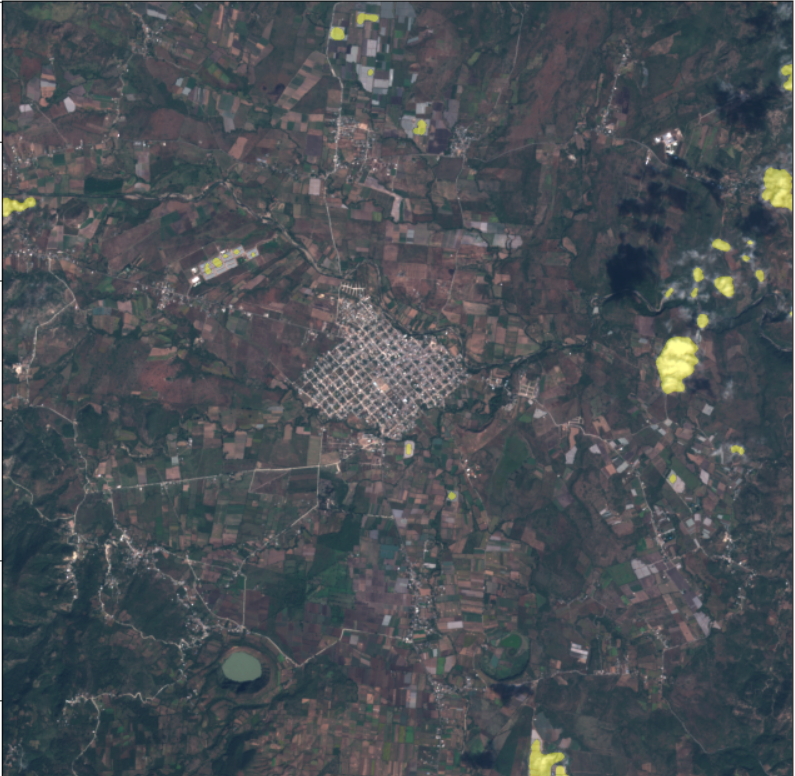}
    \caption{Sentinel2 image after cloud detection}
\end{figure}

We have also checked the detection results for only the RGB bands of the satellite image of sentinelhub. This time, we have taken a series of satellite images of different timestamps to detect the clouds. It can also identify the clouds efficiently using only the RGB bands.

\begin{figure}[htp]
    \centering
    \includegraphics[width=8cm]{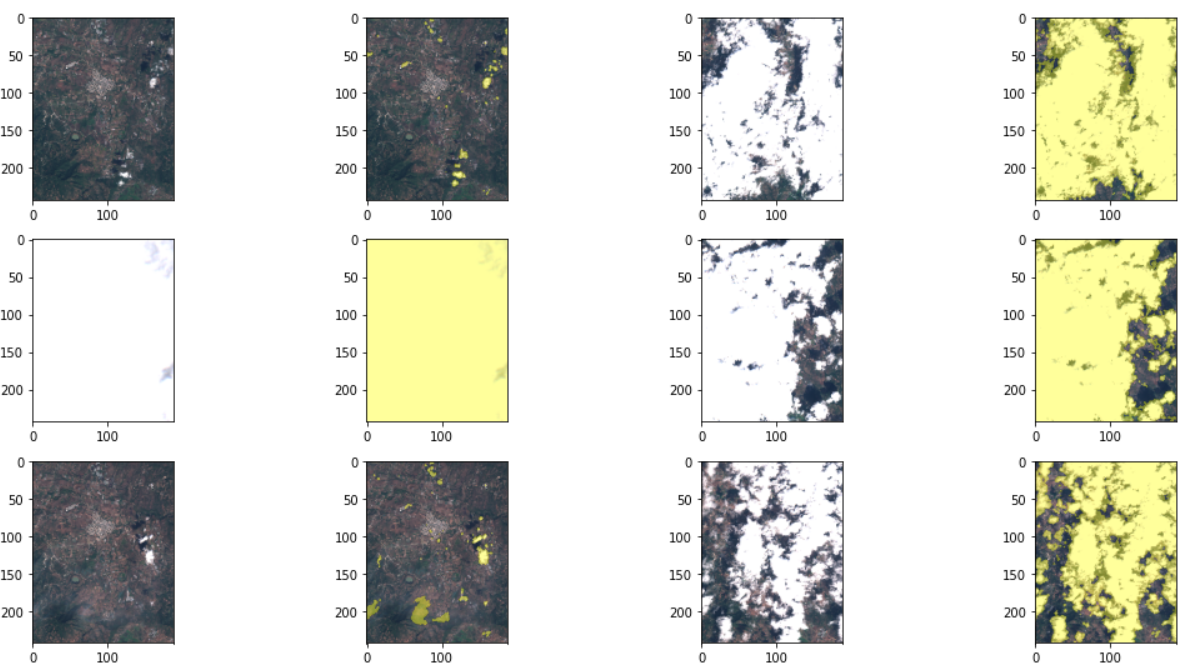}
    \caption{Sentinel2 image after cloud detection}
\end{figure}

\subsection{Cloud Removal using GAN}
We have used \cite{b7} as reference to implement cloud removal from the satellite images. In this paper, author's have proposed a spatial attention generative adversarial networks. We have taken reference of spatial attentive block and spatial attentive residual block to create the architecture of generator. The Figure 5 and Figure 6 show the block.

\begin{figure}[htp]
    \centering
    \includegraphics[width=7cm]{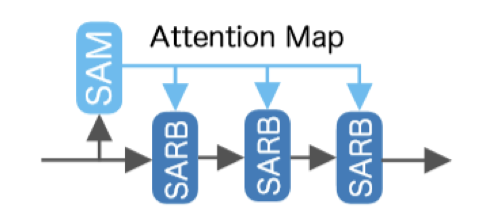}
    \caption{Spatial Attentive Block}
\end{figure}

\begin{figure}[htp]
    \centering
    \includegraphics[width=8cm]{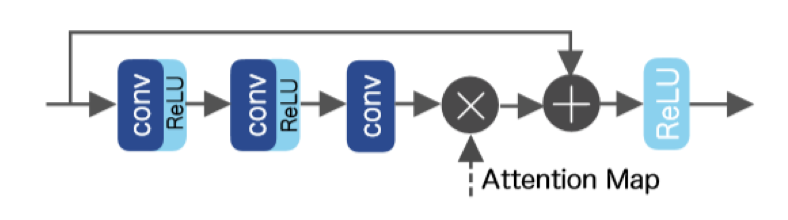}
    \caption{Spatial Attentive Residual Block}
\end{figure}

We have implemented 2 new approaches to improve results of cloud removal using spatial attention generative adversarial network. In the first approach we have changed the architecture of the generator network. This new architecture first extracts the features from image and then generate spatial attention map. This generated spatial map is given as input to extract features and generate new spatial map from that features and in the last given to convolution layer, which give reconstructed cloud-free image as output. Using this architecture, in the result, the PSNR is slightly increased and SSIM is similar to previous solution. For the second approach, we have used 8-neighbour pixels to generate spatial attention map. This approach shown a significant increase in PSNR and slight improvement in SSIM. Using 8-neighbour it has made significant change in the genrated attention map which resulted in better reconstruction of cloud-free images from cloudy images.

To train the GAN model, we have used 400 images from RICE dataset from which 320 images are used to train the model and 80 images are used to validate the model. We trained the model for 100 epochs. Table 1 shows the comparison between 3 solutions to remove clouds from satellite images in terms of PSNR and SSIM. Figure 7, Figure 8 and Figure 9 shows comparison between result of previous solution and our proposed solution.
\begin{table}
    \centering
    \begin{tabular}{|c|c|c|}
         \hline
         & PSNR & SSIM \\
         \hline
        Previous Solution & 26.4239 & 0.9545 \\
        \hline
        Our Solution  &  29.4300 & 0.9670  \\
        \hline
    \end{tabular}
    \caption{Comparison of PSNR \& SSIM}
    \label{tab:my_label}
\end{table}

\begin{figure}[htp]
    \centering
    \includegraphics[width=3cm]{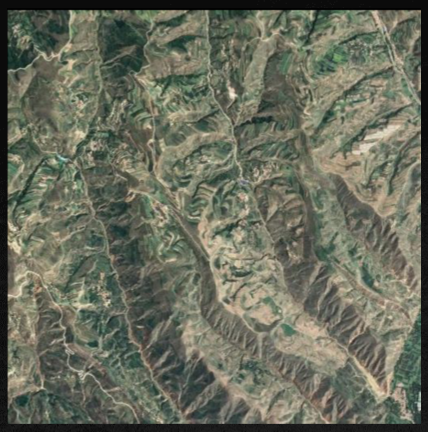}
    \caption{Ground Truth}
\end{figure}

\begin{figure}[htp]
    \centering
    \includegraphics[width=3cm]{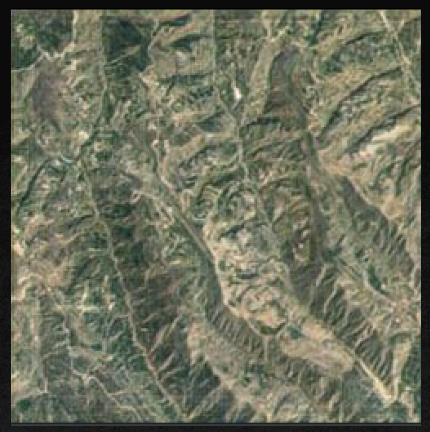}
    \caption{Our solution}
\end{figure}

\begin{figure}[!htp]
    \centering
    \includegraphics[width=3cm]{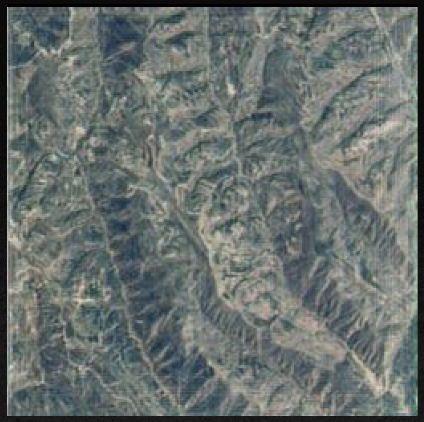}
    \caption{Previous Solution}
\end{figure}

\section{Future Work and Conclusion}
In this project, we have implemented previously available cloud detection technique using sentinel hub tool and cloud removal techniques using GAN and tried to improve the available algorithms. Using GAN, we have successfully generated the cloud-free images from cloudy satellite images and improved the reconstructed image quality. 
For the future work, we can improve the model architecture using U-net like architecture and integrating spatial and temporal information available of satellite images by training the model using multiple cloudy images.

\end{document}